\documentclass[11pt,twocolumn]{scrartcl}

\usepackage{casa_conf}	
\usepackage{graphicx}	
\usepackage{amsmath}
\usepackage{algorithm}
\usepackage{algorithmic}
\usepackage{booktabs}
\usepackage{multirow}

\title{Emotion Loss Attacking : Adversarial Attack Perception for Skeleton based on Multi-dimensional Features}

\author{Feng Liu\\
       East China Normal University\\
       lsttoy@163.com\\
       \and
       Qing Xu\\
       Beijing University of Posts and Telecommunications\\
       xqing0908@163.com\\
       \and
       Qijian Zheng\\
       East China Normal University\\
       shange0403@163.com\\
       }

\begin{document}

\maketitle

\begin{abstract}
Adversarial attack on skeletal motion is a hot topic. However, existing researches only consider part of dynamic features when measuring distance between skeleton graph sequences, which results in poor imperceptibility. To this end, we propose a novel adversarial attack method to attack action recognizers for skeletal motions. Firstly, our method systematically proposes a dynamic distance function to measure the difference between skeletal motions. Meanwhile, we innovatively introduce emotional features for complementary information. In addition, we use Alternating Direction Method of Multipliers(ADMM) to solve the constrained optimization problem, which generates adversarial samples with better imperceptibility to deceive the classifiers. Experiments show that our method is effective on multiple action classifiers and datasets. When the perturbation magnitude measured by l norms is the same, the dynamic perturbations generated by our method are much lower than that of other methods. What's more, we are the first to prove the effectiveness of emotional features, and provide a new idea for measuring the distance between skeletal motions.
\end{abstract}
\linebreak
\linebreak


\section{Introduction}
Affective computing\cite{picard2000affective} is one of the hotspots in today's AI research, which includes and is not limited to the research of facial emotion recognition\cite{zhang2021off}, speech emotion recognition\cite{liu2022lgcct}\cite{10095193}, gesture emotion recognition\cite{xu2022aes}, multimodal emotion recognition\cite{fu2024lmr}\cite{liu2023tacfn} and some personality recognition\cite{liu2022opo} based on dynamic expression recognition\cite{Wang_2023_CVPR} and other related technologies\cite{liu2022evogan}.
As the research of skeleton-based action recognition is becoming more and more popular, its robustness in practical application scenarios has attracted extensive attention and exploration\cite{liu2020adversarial,zheng2020towards,bai2020robustness}. Researches in adversarial attack has revealed that deep learning methods are vulnerable to carefully devised data perturbations. Adversarial attacks on static data such as images and text have been widely studied\cite{xiao2019meshadv}, but the research on time-series data especially skeleton data is relatively lack and immature\cite{karim2020adversarial,fawaz2019adversarial,jia2020robust}. The adversarial attack on skeleton sequences mainly faces two challenges: low redundancy and perceptual sensitivity, which is unique from static data and other time-series data. 
A skeletal motion is composed of joints and physical relationships among joints. That means the action domain of skeletal motions is limited, and the requirements of imperceptibility for adversarial samples become more stringent. A skeletal motion usually has less than 100 Degrees of freedom (Dofs), much lower than images/meshes. What's more, any sparsity based perturbation (on a single joint or a single frame) will greatly affect the dynamics (leading to jitter or bone-length violation) and is very obvious to the observer. 

There are many typical attack methods for classified networks which can be divided into gradient based and optimization based. These methods are mostly based on images\cite{xiao2018generating}that are also called European structure. However, skeletal motion is with non European structure. That means the distance measurement of skeletal motion can not be simply measured by l0, l2 and others. However, the existing researches\cite{diao2021basar,wang2021understanding} don't consider it and only use perturbations magnitude as the metric to judge the imperceptibility\cite{wang2021understanding}, which is unreasonable for skeletal motions. Therefore, we take the unique dynamics into consideration comprehensively when generating adversarial samples.


Research on emotion recognition is becoming more and more popular, such as speech-based emotion recognition\cite{Morais2021SpeechER}, text-based emotion recognition\cite{li2021fusing}, multimodal emotion recognition\cite{liu2022group,fu2021lmr} and action-based emotion recognition\cite{hu2021tntc}. Previous studies have revealed that emotion can be reflected from action. Moreover, the dynamic features mentioned earlier can measure the visual difference between the two samples, and the emotional features can reveal the logical relationship between skeleton joints. Therefore, we try to introduce emotional feature as one of the indicators to measure the difference between samples, and explore the impact of emotional features.

To this end, in this paper we propose an adversarial attack method for skeletal motions. Firstly, we define a novel distance function based on multi-dimensional features that considers dynamics and introduce emotional features for the first time. The distance considers spatial dynamic such as bone length and bone angle and temporal dynamic like speed. In addition, we use Alternating Direction Method of Multipliers(ADMM) to ensure the imperceptibility. Our proposed attack method is evaluated on three kinds of state-of-the-art models. Extensive evaluations show that our attack method can achieve 100\% success rate with almost no violation of the constraints mentioned above. To summarize, 
the contributions of this paper are as follows:

(1) We define a novel distance method based on multi-dimensional features to measure skeletal motions, which includes dynamic distance and innovatively incorporate emotional features as complementary information. 

(2) We propose an effective optimization algorithm based on Alternating Direction Method of Multipliers(ADMM) to solve the primal constrained problem, which generates adversarial skeletal motion samples with perturbations as few as possible.

(3) We fully evaluate multiple state-of-the-art models and multiple datasets and verify that the adversarial samples generated by ours can successfully deceive models with fewer perturbations and lower imperceptibility. 


\begin{figure*}[htb]
  \centering
\begin{minipage}[c]{0.8\textwidth}
   \includegraphics[width=1.0\linewidth]{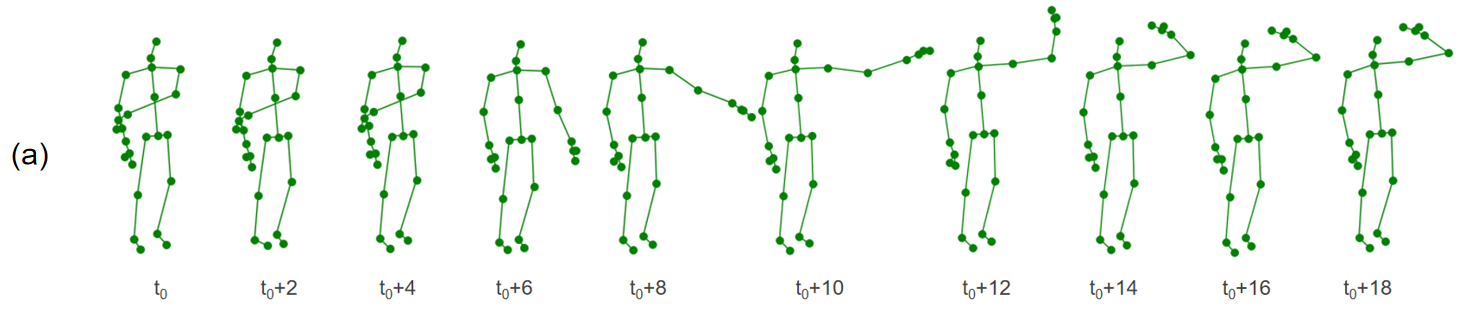}
   
  \end{minipage}
  
  \begin{minipage}[c]{0.8\textwidth}
   \includegraphics[width=1.0\linewidth]{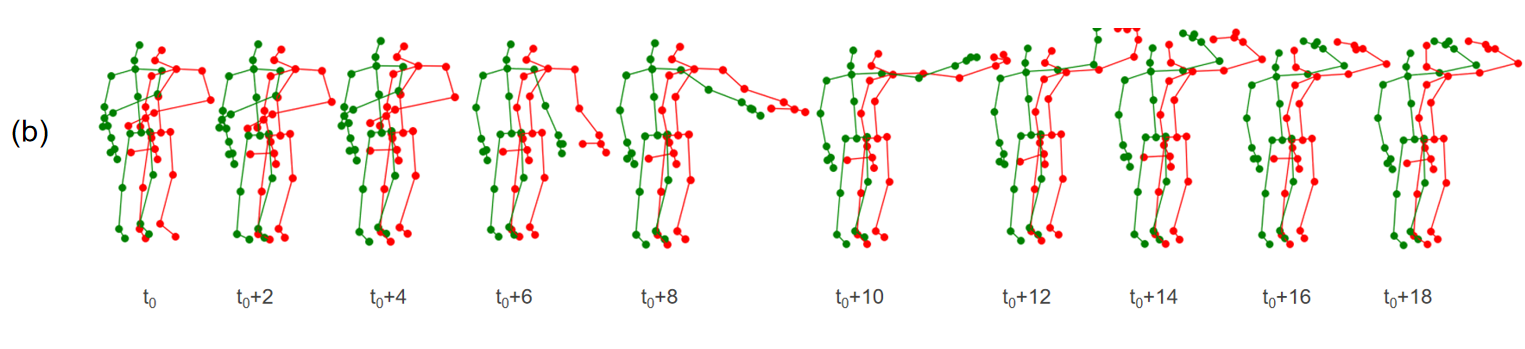}
  \end{minipage}
  
  \begin{minipage}[c]{0.8\textwidth}
   \includegraphics[width=1.0\linewidth]{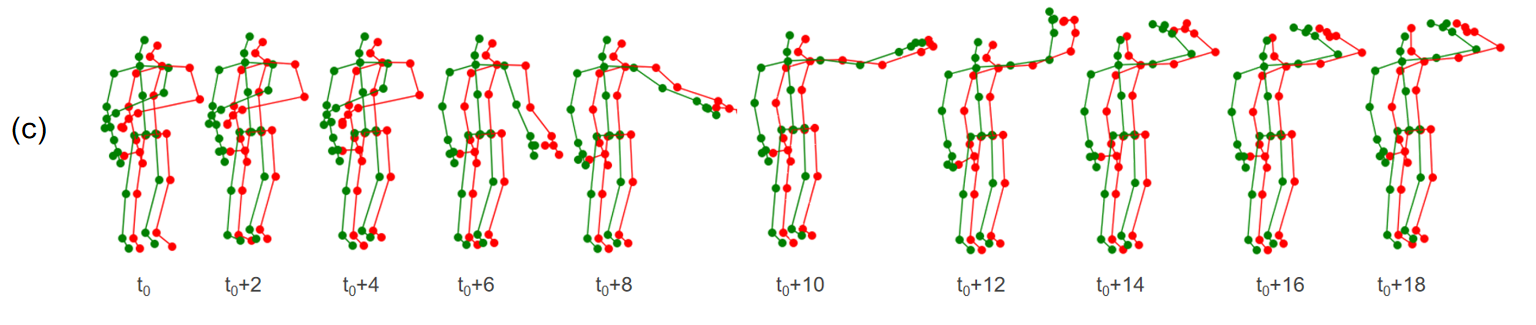}
  \end{minipage}
  
  \begin{minipage}[c]{0.8\textwidth}
   \includegraphics[width=1.0\linewidth]{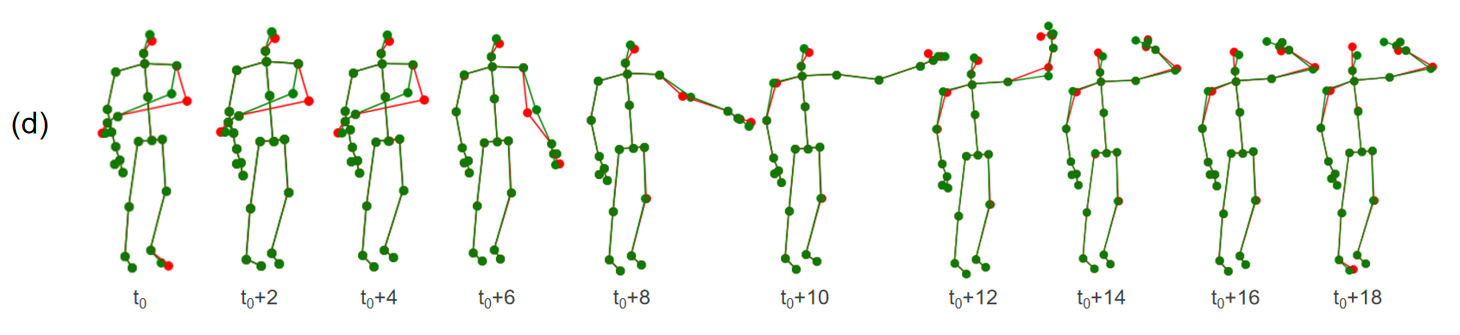}
  \end{minipage}
   
   \caption{Visual comparison. The green joints represent the original sample and the red ones represent the adversarial sample. (a) shows the original sample, (b) shows attack results of C\&W, (c) shows attack results of SMART, (d) shows attack results of our method.}
   \label{fig:perceptual}
\end{figure*}

\section{Methodology}
Given a skeleton sample $x$, $l$ is the predicted label of $x$ of a trained classifier. We denote $\Theta(x)$ as the result of the probability of each class before softmax layer and $F$ as softmax function. We aim to find minimum perturbation added to original sample to get adversarial sample $x^{'}$ and $F(\Theta(x)) \neq F(\Theta(x^{'}))$. The problem can be formulated as:
\begin{equation}
\begin{split}
    min\quad &D(x, x^{'})  \\
    subject\; to\quad &F(\Theta(x^{'})) = l^{'}, x^{'} \in [0, 1]^{n}
  \label{eq:eq1}
\end{split}
\end{equation}
where $D$ is the distance function to measure original sample and adversarial sample, $l^{'}$ is the predicted label of adversarial sample $x^{'}$ and $l^{'} \neq l $ is a hard constraint. The hard constraint of classification is defined according to different attack modes. In addition, we use emotion feature as supplementary expression of skeletal motion.

\subsection{Skeleton-based dynamic distance}

For the motion $h\{m\}=\{m_0, m_1, \cdots, m_t\}$, $m_t$ at time t consists of not only 3D coordinates of joints but the connected topological structure. A skeleton has its dynamic information and we can represent dynamics of a skeleton from spatial and temporal aspects.

From spatial perspective, we need static and dynamic information. To ensure imperceptibility bone lengths should remain the same in adversarial samples. So we use bone length as static spatial dynamic constraint. A bone corresponds to two joints so a bone can be represented as a vector $B$=($x_s-x_t$, $y_s -y_t$, $z_s - z_t$) where ($x_s,y_s,z_s$) is coordinate of source joint and ($x_t,y_t,z_t$) is target joint. In this regard, length of the $i_{th}$ bone at the $t_{th}$ frame is defined as $B_i^t$ = $\sqrt{(x_{s}-x_{t})^2 + (y_{s}-y_{t})^2 + (z_{s}-z_{t})^2}$. Normally, the length of the bone should be the same between original sample and adversarial sample. It can be represented as $b(x, x^{'}) = \lvert B_i^t - B_i^{'t} \rvert / B_i^t$ where $B_i^t$ and $B_i^{'t}$ represent the original sample and the adversarial sample respectively.

We use angle between bones as measurement of change of skeletons. Every two connected bones form an angle, and the change of angles means the extent of bone rotation. So we introduce angle constraint into skeleton dynamic distance. To avoid gradient explosion, we compute change of the $i_{th}$ angle at $t$ frame $A_i^t$ by\cite{zheng2020towards}. Similar to bone length, we use function $a(x, x^{'})$ to measure changes of angles.

From temporal perspective, it is necessary to ensure temporal smooth of adversarial samples. So we introduce joint's speed as an index of temporal measurement. We can estimate speed of the $i_{th}$ joint at frame $t$ by the Euclidean distance between two consecutive temporal frames. The speed of the $i_{th}$ joint at frame $t$ is computed as  $S_i^t$ = $\sqrt{(x_{i}^{t+1}-x_{i}^{t})^2 + (y_{i}^{t+1}-y_{i}^{t})^2 + (z_{i}^{t+1}-z_{i}^{t})^2}$. The measure is represented as $s(x, x^{'}) = \lvert S_i^t - S_i^{'t} \rvert / S_i^t$ where $S_i^t$ and $S_i^{'t}$ 
represent the joint of original sample and adversarial sample respectively. Similar to spatial constraints, $\varepsilon_s$ is maximum change value.

\begin{algorithm}
\caption{Generating adversarial samples}
\label{alg:algorithm1}


\begin{algorithmic}[1]
\STATE \textbf{Input:} original sample $x$, maximum numbers of iterations $I$, Classification Loss Function $C$, Dynamic Distance Function $D_d$, Emotion Distance Function $D_e$

\STATE \textbf{Initialization:} $x_{0}^{'}$ = $x$, Lagrangian Variable: $\lambda$

\WHILE{$i \leq I-1$}

        \STATE $x'$($i+1$)=$argmin_{x}$  $L$($x'$($i$),  $\lambda$($i$));
        
        \STATE $l_d$, $l_e$, $l_c$=$D_d$ ($x'$($i+1$)), $D_e$($x'$($i+1$)), $C$($x'$($i+1$));
        
        \STATE $l_c$ = $\lambda$ $\times$ $l_c$ + $\frac{\gamma}{2}$($\lvert \lvert l_c \rvert \rvert^2_{2})$;
        
        \STATE $loss$ = $l_d$ + $l_e$ + $l_c$;
        
        \STATE $Backward$($loss$);
        
        \STATE $\lambda$($i+1$)= $\lambda$({$i$}) + $\gamma$ $C$($x'$($i+1$));
\ENDWHILE
\end{algorithmic}
\end{algorithm}

\subsection{Classification loss}

\emph{Untargeted Setting}. In mode of untargeted attack, $F(\Theta(x^{'})) \neq l$ means that predicted label of the classifier can be any label other than the ground truth $l$. That means maximum value of possibility must not be $l$, which can be represented as $max(\Theta(x^{'})) > \Theta_{l}(x^{'})$. Based on this, We can denote classification loss as  $max(\Theta(x^{'})) - \Theta_{l}(x^{'}) > conf$ where $conf$ is expected value of wrong prediction of classifiers. Note that the inequality constraints in the original problem impose inequality constraints on the corresponding Lagrangian variables in the dual problem. So we convert inequality constraint to equality constraint as:
\begin{equation}
    max(\Theta_{l}(x^{'}) - max(\Theta(x^{'})) + conf) = 0 
  \label{eq:classify_un}
\end{equation}

\emph{Targeted Setting}. Under the setting of targeted attack, we aim to get $F(\Theta(x^{'})) = l_{t}$. That is to say, $max(\Theta(x^{'})) = \Theta_{l_{t}}(x^{'})$. So we can use $\Theta_{l_{t}}(x^{'}) - max_{l \neq l_{t}}(\Theta(x_{'})) > conf$ as classification loss. The equation form is expressed as follows:
\begin{equation}
    max(\Theta(x^{'}))  -\Theta_{l_{t}}(x^{'}) = 0 
  \label{eq:classify}
\end{equation}

\subsection{Emotion loss}

In addition to dynamics, skeletal motions may also contain emotional features. Therefore, we innovatively introduce emotions as non-dynamic features to measure the distance. Specifically, we use the emotion classifier in literature\cite{narayanan2020proxemo} to extract the emotional features of skeletal motions. The model embeds the skeletal motion sequence into images for training, and obtains the predicted emotional features through four group convolution. Group convolution allows the network to learn independently from different parts of the input, so as to determine the joint interval that has the greatest impact on the final category. We put gait-based skeletal motions and the generated adversarial samples into the pre-training model to obtain the distance loss of emotion in non-dynamic features. We use $E$ as emotional features. The specific formula is $e(x, x^{'}) = \lvert \lvert E(x) - E(x^{'}) \rvert \rvert$.

\subsection{Optimal dual method}

The objective function of the constrained optimization problem formulated as equation~\ref{eq:eq1}. $D(x, x^{'})$ is distance function and $D(x, x^{'})$ = $b(x, x^{'}) + a(x, x^{'}) + s(x, x^{'}) + e(x, x^{'})$. The hard constraint is loss of classification as denoted in equation~\ref{eq:classify_un} and equation~\ref{eq:classify}. In optimization theory, the optimization problem of objective function under constraints can be transformed into a corresponding dual problem. Due to its strong duality, the solution of the original problem can be obtained by solving the dual problem. We introduce the Alternating Direction Method of Multiplier to solve the dual problem.
We denote Lagrange expression as $L(x, \lambda) = b(x, x^{'}) + a(x, x^{'}) + s(x, x^{'}) + e(x, x^{'}) + \lambda C(l, l^{'}) + \frac{\gamma}{2}  \lvert \lvert C(l, l^{'}) \rvert \rvert _{2}^{2}$ where $\lambda$ is Lagrange multiplier and $C$ is classification loss. In order to effectively find the local optimal solution, we use Adam optimization algorithm for Adam optimization algorithm always converges faster than vanilla SGD. The process of generating adversarial samples is described as Algorithm~\ref{alg:algorithm1}.

\begin{table*}
  \centering
    \caption{The results of our method with untargeted attack mode on NTU RGB+D.}
  \resizebox{\textwidth}{!}{
  \begin{tabular}{lcccccc|ccccccccccccc}
    \toprule
    \multirow{2}*{Models} & \multirow{2}*{$\gamma$} & \multicolumn{5}{c|}{NTU RGB+D CV} & \multicolumn{5}{c}{NTU RGB+D CS}\\
    &  & $\triangle$B/B & $\triangle$A/A &  $\triangle$S/S & SR & l2 & $\triangle$B/B & $\triangle$A/A &  $\triangle$S/S & SR & l2\\
    \hline
    \multirow{3}*{HCN} & 0.1 & 0.9\% & 4.2\%  & 3.2\% & 100\% & 0.26 & 0.8\% & 4.1\%  & 3.3\% & 100\% & 0.24\\
      & 1.0 & 1.3\% & 6.7\%  & 4.2\% & 100\% & 0.21 & 1.3\% & 6.7\%  & 4.2\% & 100\% & 0.21\\
      & 10.0 & 2.4\% & 15.1\%  & 7.5\% & 100\% & 0.22 & 2.3\% & 13.7\%  & 7.1\% & 100\% & 0.20\\
    \hline
    \multirow{3}*{2s AGCN} & 0.1 & 0.4\% & 2.5\%  & 1.8\% & 100\% & 0.07 & 0.6\% & 2.8\%  & 1.6\% & 100\% & 0.06\\
     & 1.0 & 0.6\% & 2.7\%  & 2.2\% & 100\% & 0.08 & 0.5\% & 2.1\%  & 1.9\% & 100\% & 0.07\\
     & 10.0 & 2.3\% & 12.0\%  & 6.9\% & 100\% & 0.15 & 1.4\% & 7.8\%  & 4.1\% & 100\% & 0.12\\
    \hline
    \multirow{3}*{SGN} & 0.1 & 0.6\% & 3.1\%  & 2.1\% & 100\% & 0.15 & 0.8\% & 3.0\%  & 1.8\% & 100\% & 0.14\\
    
     & 1.0 & 0.9\% & 4.8\%  & 2.4\% & 100\% & 0.12 & 0.9\% & 4.9\%  & 1.7\% & 100\% & 0.12\\
     
     & 10.0 & 2.6\% & 12.6\%  & 7.5\% & 100\% & 0.18 & 1.8\% & 10.6\%  & 4.5\% & 100\% & 0.13\\
    
    \bottomrule
  \end{tabular}}
  \label{tab:result1}
\end{table*}
\section{Experiments}
\subsection{Models and datasets}
\textbf{NTU RGB+D} consists of 25 joint points in each skeleton. The original paper recommends two benchmarks: (1) Cross-subject: the subject in training set and validation set are different. (2) Cross-view: training set captured by camera 2 and 3 and validation set captured by camera 1. \textbf{Kinetics-400}\cite{kay2017kinetics} contains 18 joints in each skeleton. The adversarial samples of following experiments are generated on the validation set on two datasets.

We select HCN\cite{li2018co}, 2s AGCN\cite{shi2019two}, SGN\cite{zhang2020semantics} and investigate their vulnerability under different scenarios. HCN\cite{li2018co} has achieved state-of-the-art performance before GCN related work and 2s AGCN\cite{shi2019two} is an effective GCN-based model. SGN\cite{zhang2020semantics} introduces semantics for the first time and achieves great performance.

\subsection{Evaluation metrics}
For the adversarial attack method, the effectiveness refers to the extent to which the method can provide "successful" adversarial samples. On this premise, we evaluates the quality of adversarial samples from two aspects: misclassification and imperceptibility.

(1) \textit{Misclassification} refers to the degree of deception of adversarial samples, reflected in the following indicators.

\textbf{Attack Success Rate(SR)}, that is, the proportion of adversarial samples wrongly classified (in untarget mode) or wrongly classified to the specified class (in target mode). $SR_{UA} = \frac{1}{N}\sum_{j=0}^{N} sum(F(x)\neq l_{t})$ is under untarget mode and $SR_{TA} = \frac{1}{N}\sum_{j=0}^{N} sum(F(x)=l_{t})$ is under target mode.



(2) \textit{Imperceptibility}. We defined four evaluation metrics for the original sample $x$ and adversarial sample $x^{'}$: the average deviation percentage of bone length $\triangle \textbf{B/B} =\frac{\sum_{i=0}^{N}(\sum_{j=0}^{M}(B_{j}^{i}-B_{j}^{'i})/B_{j}^{i})}{N \times M}$, the average deviation percentage of bone angle $\triangle \textbf{A/A}$, the deviation percentage of joint speed $\triangle \textbf{S/S}=\frac{\sum_{j=0}^{N}\lvert \lvert x_{s}-x_{s}^{'}\rvert\rvert_{2}}{F\times N \times O}$ and the l2 distance between the original sample and adversarial sample $\textbf{l2}=\frac{\sum_{j=0}^{N}\lvert \lvert x-x^{'}\rvert\rvert_{2}}{F\times N}$ 
where $N$ is the total number of adversarial samples, $F$ is the total number of frames in a motion and $O$ and $M$ are the total number of joints and frames in a skeleton.



\subsection{Attack results}
\textit{\textbf{Misclassification}}. 
The quantitative results of untargeted attack mode are shown in Table~\ref{tab:result1}. Our method achieves high success rates across different datasets and target models. For targeted attack mode, results are shown in Table~\ref{tab:targeted}. It is not surprising to turn 'reading' into 'writing'. Therefore we choose 'drinking water' as targeted skeletal motion class for obvious differences are existed. For classifier model 2s AGCN and SGN, we can see that the adversarial samples successfully deceives the two models but more perturbations are added to the adversarial samples to deceive SGN compared with 2s AGCN. This shows that even though the network of SGN is relatively simple, its defense capability exceeds that of 2s AGCN. What's more, it is also proved that semantics can greatly enhance the ability of network to learn logical and dynamic features of skeletal motions. 
\begin{table}[htbp]
  \centering
    \caption{The results of our method on targeted attack mode on NTU RGB+D with cross view setting.}
  \resizebox{0.5\textwidth}{!}{
  \begin{tabular}{lccccccccccccccccccc}
    \toprule
       & $\triangle$B/B & $\triangle$A/A &   $\triangle$S/S & SR & l2\\
    \hline
    HCN & 3.1\% & 15.4\% & 7.5\% & 100\% & 0.62\\
    2s AGCN & 1.1\% & 4.8\% & 3.1\% & 100\% & 0.21\\
    SGN & 2.9\% & 15\% & 4.2\% & 100\% & 0.44\\
    \textbf{AeS-GCN\cite{xu2022aes}} & \textbf{10.5}\% & \textbf{41.2}\% & \textbf{21.9}\% & \textbf{73.9}\% & \textbf{5.92}\\
    \bottomrule
  \end{tabular}}
  \label{tab:targeted}
\end{table}

\textit{\textbf{Imperceptibility}}.
Our method obtains adversarial samples with much lower perturbation as shown in Table~\ref{tab:cw_results}. Therefore, it is also proved that the dynamic distance proposed is more effective than l2 distance for skeletal motions. Compared with the existing methods, strict perceptual control is used as the optimization target problem to improve the imperceptibility. Taking the adversarial sample generated based on SGN as an example in Figure~\ref{fig:perceptual}, we sample 10 frames from sequence for visual display. Since the previous results show that SGN produces greater perturbations when it is successfully attacked. The label of original sample is 'throwing' and after attack the predicted label is 'brushing teeth' under targeted attack mode. When comparing two samples carefully, we can find that differences are existed in some joints. However, when two samples are played as video sequences, differences are hard to find. In addition, we also find that the perturbations added to adversarial samples are concentrated on arms and hands, which contain obvious dynamics. This may remind us that we should consider to reduce sensitivity to special data when designing classifiers.

\begin{table}[htbp]
  \centering
  \caption{The effectiveness of emotional features on HCN on NTU RGB+D with cross view setting.}
    \resizebox{0.5\textwidth}{!}{
  \begin{tabular}{ccccccccc}
  
    \toprule
         & $\triangle$B/B & $\triangle$A/A & $\triangle$S/S & SR & l2\\
    \hline
    Ours w emotion & 1.3\% & 6.7\% & 4.2\% & 100\% & 0.21\\
    Ours w/o emotion & 1.2\% & 7.1\% & 4.8\% & 100\%  & 0.24\\
    
    \bottomrule
  \end{tabular}}
  \label{tab:emo_results}
\end{table}
We also verify the role of emotional features. Table~\ref{tab:emo_results} shows the effectiveness of emotional features. Emotional features can control the perception of adversarial samples from overall perspective. However, the improvement is not obvious, which may be due to the accuracy of the emotion recognizer used. Therefore, it can be expected that with in-depth study of emotion recognition, emotional features will be better integrated.


\begin{table}
  \centering
  \caption{The results of C\&W on HCN on NTU RGB+D.}
  
  \begin{tabular}{p{1.4cm}p{1.4cm}p{0.8cm}p{0.8cm}p{0.8cm}cccccc}
    \toprule
      &   & $\triangle$B/B & $\triangle$A/A &  SR & l2\\
    \hline
    \multirow{2}*{Untargeted} & NTU CV & 4.67\% & 24.1\% & 100\% & 0.28\\
    & NTU CS & 4.09\% & 21.1\%& 100\%  & 0.24\\
    \hline
    \multirow{2}*{Targeted} & NTU CV & 8.8\% & 46.8\% & 100\% & 0.51\\
    & NTU CS & 9.5\% & 50.7\%& 100\%  & 0.52\\
    \bottomrule
  \end{tabular}
  \label{tab:cw_results}
\end{table}


\section{Conclusion}
To summarize, in order to explore the vulnerability of skeleton-based action recognizers, we propose a novel attack method for based on multi-dimensional features. We fuse dynamics and emotional features of skeletal motions and generate successful adversarial samples based on ADMM. A large number of experiments show that our method has fewer perturbations and better imperceptibility than other methods. Our method is effective on multiple datasets and the state-of-the-art models. In the future, we will systematically study how to improve the defense ability of models to resist attacks.

\section*{Acknowledgements}
This work supported by Beijing Key Laboratory of Behavior and Mental Health in School of Psychological and Cognitive Sciences, Peking University. We would also like to thank the colleagues in the CiL lab at East China Normal University for their efforts on this project, and the reviewers for their time and hard work.


\bibliographystyle{unsrt}
\bibliography{refs}

\end{document}